\documentclass[final]{cvpr}
\usepackage[T1]{fontenc}

\usepackage{times}
\usepackage{epsfig}
\usepackage{graphicx}
\usepackage{amsmath}
\usepackage{amssymb}

\usepackage{amsfonts}%
\usepackage{nicefrac}%
\usepackage{xcolor}
\usepackage{comment}
\usepackage{siunitx}
\usepackage{stfloats} %
\usepackage[pagebackref=true,breaklinks=true,colorlinks,bookmarks=false]{hyperref}

\def\confYear{Arxiv 2022}
\makeatletter
\renewcommand{\paragraph}{%
\@startsection{paragraph}{4}{\z@}%
{1.25ex \@plus 1ex \@minus .2ex}{-1em}%
{\normalfont\normalsize\bfseries}%
}
\makeatother

\newcommand{\rodrigo}[1]{}
\newcommand{\jordi}[1]{}
\newcommand{\vitto}[1]{}

\begin{document}

\title{From colouring-in to pointillism:\\revisiting semantic segmentation supervision}

\author{%
Rodrigo Benenson \\
Google Research \\
  \texttt{benenson@google.com} \\
  \and
Vittorio Ferrari \\
Google Research \\
\texttt{vittoferrari@google.com} \\
}

\maketitle

\begin{abstract}
The prevailing paradigm for producing semantic segmentation data relies
on densely labelling each pixel of each image in the dataset, akin
to colouring-in books. This approach becomes a bottleneck when scaling
up in the number of images, classes, and annotators.
Here we propose instead a
pointillist approach for semantic segmentation annotation, where
only point-wise yes/no questions are answered.
We explore design alternatives for such an active learning approach, measure the speed and consistency of human annotators on this task, show that this strategy enables training good segmentation models, and that it is suitable for evaluating models at test time.
As concrete proof of the scalability of our method, we collected and released 22.6M point labels over 4,171 classes on the \href{https://g.co/dataset/open-images}{Open Images dataset}.
Our results enable to rethink the semantic segmentation pipeline of annotation,
training, and evaluation from a pointillism point of view.
\end{abstract}

\begin{figure*}
\begin{centering}
\includegraphics[height=0.5\columnwidth]{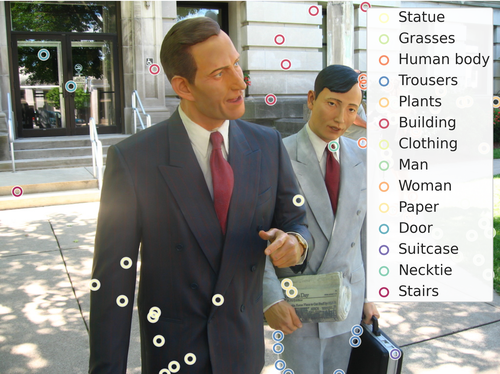}\hfill%
\includegraphics[height=0.5\columnwidth]{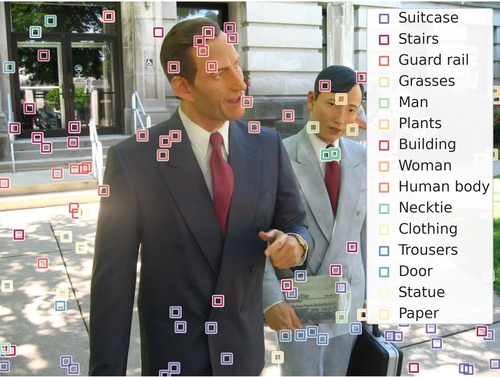}%
\hfill%
\includegraphics[height=0.5\columnwidth]{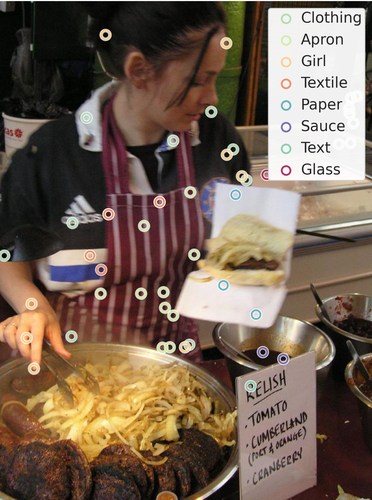}\hfill%
\includegraphics[height=0.5\columnwidth]{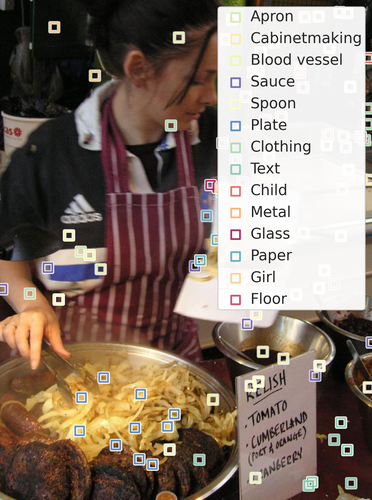}
\caption{\label{fig:yes-no-examples}Example of collected yes and no point labels (circles and squares respectively). Additional examples in appendix \S\ref{sec:example_yes_no_points}.}
\end{centering}
\end{figure*}%

\section{Introduction}
\label{sec:Introduction}

As computer vision applications expand, so does the need for supervised
training data. Through the years, solutions for object recognition
evolved towards increased spatial accuracy from image-level class
labels~\cite{feifei04cvpr,russakovsky15ijcv,kuznetsova20ijcv}, to bounding boxes~\cite{everingham15ijcv, lin14eccv, kuznetsova20ijcv}, to semantic segmentation ~\cite{everingham15ijcv, caesar18cvpr}, to instances
segmentation~\cite{lin14eccv, cordts16cvpr, gupta19cvpr, benenson19cvpr}.
Each level requires increasing manual annotation effort. %

It is believed that humans handle about \num{30000} visual nouns~\cite{smith14multimedia, biederman87pr, deng10eccv}.
Image-level annotations covering tens of thousands of different
classes have been collected~\cite{kuznetsova20ijcv,russakovsky15ijcv}.
However at the segmentation level existing datasets typically include only a few hundred
classes~\cite{zhou19ijcv,benenson19cvpr,gupta19cvpr}. 
The annotation effort required for creating new datasets for semantic segmentation
has become a major hurdle to overcome for new applications.
Up to now segmentation annotations have resembled colouring-in
books, where the annotators provide the right colour for each pixel,
and spread this colour densely making sure to respect the class boundaries
adequately~\cite{russell08ijcv, cordts16cvpr, caesar18cvpr,zhou19ijcv}.
This is time consuming and generates many redundantly annotated pixels.

In this paper we propose an annotation strategy for semantic segmentation that can scale to thousands of classes. Starting from given image-level labels, the annotators answer only point-wise yes/no questions automatically generated by a computer model (thus “pointillism”, in reference to the late 19th century painting style).
This approach features extremely simple questions which the annotators understand immediately and can answer rapidly, without undergoing any training nor having any understanding of the notion of object boundaries or segmentation masks. This enables to access a larger pool of annotators. Other advantages include easy parallelization, enabling incremental refinement of existing annotations, and being overall more time-efficient than traditional polygon drawing tools.
The combined effect of these benefits enables scaling up segmentation annotations to many more classes, annotators, and images.
Our experiments 
show that this strategy enables training good segmentation models (%
\S\ref{sec:Training-simulations}%
),
show that it is suitable for evaluating models at test time (\S\ref{sec:Evaluating-models}),
and measure the speed and accuracy of human annotators on this task (\S\ref{sec:human-annotators}).
Finally, we run a large-scale annotation campaign (\S\ref{sec:oidv7-point-annotations}), collecting \num{22.6}M point labels over \num{4171} classes on the Open Images dataset~\cite{kuznetsova20ijcv}. These points are released as part of the point-labels mix of Open Images V7 (which totals \num{66}M point-label annotations over \num{5827} classes and \num{1.4}M images, making it one of the datasets with the most diverse range of classes with annotations suitable for semantic segmentation).
These results show that semantic segmentation is possible without dense pixel-level labelling at any stage of the pipeline. We thus argue that it is time to shift the annotations paradigm from colouring-in books towards pointillism.

\paragraph{Related work.}
\label{subsec:Related-work}
The design space of annotation strategies is large~\cite{kovashka16nwat,welinder10nips}.
Input modalities such as mouse~\cite{papadopoulos17iccv},
touch~\cite{zimmermann2012vigta}, voice~\cite{gygli19ijcv,vasudevan17cvpr},
and gaze~\cite{vaidyanathan18acl}, have been explored. Direct label
annotations are common, but pair-wise comparisons/ranking have also
been considered~\cite{gomes11nips,parikh11iccv,chen16nips_depth}.

For semantic segmentation multiple forms of direct pixel-level supervision
have been considered: vanilla dense labelling (colouring-in book)
\cite{everingham15ijcv,caesar18cvpr,zhou19ijcv,benenson19cvpr,gupta19cvpr},
dense labelling over spatial blocks~\cite{lin19iccv}, sparse points on the object interior~\cite{bearman16eccv},
boundary points~\cite{papadopoulos17iccv,maninis18cvpr},
corrective points~\cite{xu16cvpr,benenson19cvpr},
scribbles~\cite{batra2010cvpr,bearman16eccv,lin16cvpr_scribblesup}, corrective scribbles~\cite{FSS1000},
polygons~\cite{russell08ijcv,bell13tog,acuna18cvpr}, and coarse annotations~\cite{cordts16cvpr,zlateski18cvpr}.
Beyond direct supervision (possibly noisy or incomplete),
weakly supervised methods have been devised to derive approximate segmentations
starting from image-level labels~\cite{pathak15iccv,papandreou15iccv,bearman16eccv,kolesnikov16eccv,oh17cvpr,ahn19cvpr,wang20cvpr}
or bounding box~\cite{cheng15cgf,dai15iccv,papandreou15iccv,khoreva17cvpr} 
annotations.

When annotating at scale, most works do multiple passes
over the images. The COCO~\cite{lin14eccv} annotation
process first marks the classes present in the image (image-level
labels), then does instance spotting for each class, and finally segments instances separately.
Efficiently collecting image-level labels is by itself challenging.
COCO and ImageNet adopt a hierarchical approach~\cite{deng14chi},
Open Images uses human verification of machine suggestions~\cite{kuznetsova20ijcv},
LVIS uses free-form text input with auto-complete~\cite{gupta19cvpr}.
Other strategies include gamification~\cite{ahn04chi} or using voice plus mouse~\cite{gygli19ijcv,ponttuset20eccv} (implicitly naming classes and their approximate location).
In the same spirit, we assume image-level labels have been produced beforehand, and focus on moving from image-level to pixel-level annotations.

Active learning~\cite{cohn94ml, settles2009active_learning} is popular for image classification~\cite{joshi09cvpr, sener18iclr,li13cvpr,beluch18cvpr}, but only few works have
used it for semantic segmentation, either automatically selecting the
points to annotate~\cite{vezhnevets12cvpr2,sinha19iccv,shin2021cvpr}, the regions~\cite{mackowiak18bmvc} or the images~\cite{jain16cvpr,yang2017miccai}. These works however do not consider data collection with real human annotators.

Beyond the training data, it is necessary to also collect evaluation
data. This is typically done by dense labelling the test set,
and using metrics such mean intersection-over-union (mIoU, also known
as \textquotedbl Jaccard Index\textquotedbl )~\cite{everingham15ijcv},
size sensitive IoU~\cite{cordts16cvpr}, or boundary-sensitive metrics~\cite{martin04pami,csurka13bmvc,ponttuset15pami}.%

Closest to our work are~\cite{bearman16eccv,shin2021cvpr,cheng2022cvpr} which also advocate point annotations for segmentation. %
In~\cite{bearman16eccv} the annotator clicks on an object to assign a class label, whereas in our work the point location is determined by the machine, and we ask the annotator a simpler yes/no question. Compared to~\cite{bearman16eccv,shin2021cvpr,cheng2022cvpr} we additionally show that sparse points are suitable for test time evaluation (vs only for training).
Finally, we run a large-scale annotation campaign, collecting \num{22.6}M point labels (fig.~\ref{fig:yes-no-examples}) over \num{4171} classes on the Open Images dataset~\cite{kuznetsova20ijcv} (vs $72$k points over $20$ classes on the smaller PASCAL VOC~\cite{pascal-voc-2012} in~\cite{bearman16eccv}, and no sizeable data collection in \cite{shin2021cvpr,cheng2022cvpr}).

There exists a few methods dedicated to training segmentation models from sparse point labels \cite{laradji2019arxiv, mcever2020arxiv, shin2021cvpr, tang2022eccv, cheng2022cvpr}.
However, their results have been demonstrated using simulated annotations based on existing densely annotated datasets. The real data we release here could be fed into these methods to fully realize their potential.

\section{From colouring-in to pointillism}
\label{sec:colorbooks-to-points}

\begin{figure*}
\begin{minipage}[t]{0.3\textwidth}%
\begin{center}
\hspace{-1em}
\includegraphics[width=0.9\textwidth]{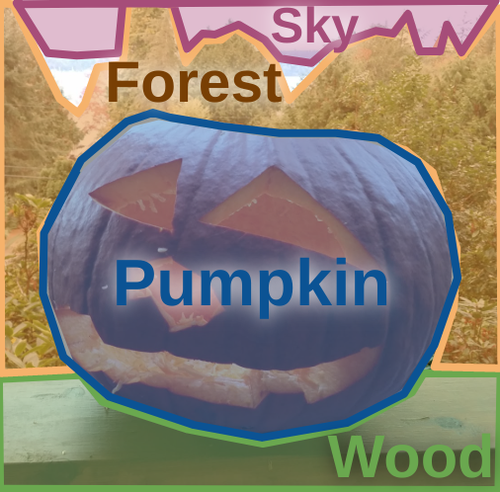}%
\caption{\label{fig:Example-of-colourbook}Example colouring-in annotations.}
\par\end{center}%
\end{minipage}\hspace*{\fill}%
\begin{minipage}[t]{0.3\textwidth}%
\begin{center}
\hspace{-2em}
\includegraphics[width=0.9\textwidth]{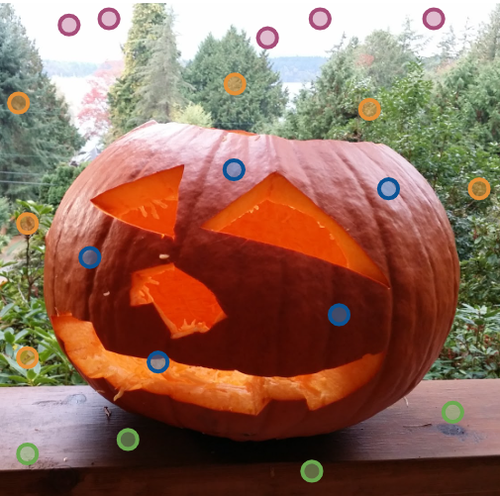}\vspace{-1.0em}
\par\end{center}
\caption{\label{fig:Example-of-pointillism}Example pointillism annotations.}
\end{minipage}\hspace*{\fill}%
\begin{minipage}[t]{0.185\textwidth}%
\begin{center}
\includegraphics[width=0.9\textwidth]{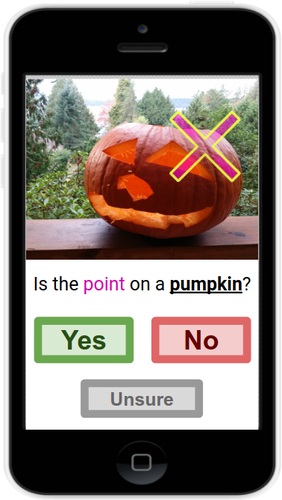}\vspace{-1.0em}
\par\end{center}
\caption{\label{fig:YesNo-UI}User interface example.}
\end{minipage}
\end{figure*}

As the field of semantic segmentation progresses, its goals become
more ambitious. The last two decades have seen dramatic
progress in the models used to tackle the problem (from simple classifiers
over handcrafted features~\cite{shotton09ijcv}
to large convolutional networks~\cite{pohlen17cvpr,chen18eccv}).
However, the methods to annotate data have evolved less.
The main strategy remains dense manual labelling either using a brush~\cite{shotton06eccv,hariharan11iccv,everingham15ijcv,caesar18cvpr}
or a polygon drawing tool~\cite{russell08ijcv,bell13tog,lin14eccv,cordts16cvpr,gupta19cvpr}.
These annotations can take 5~\cite{everingham15ijcv,bearman16eccv},
10~\cite{lin14eccv,caesar18cvpr}, or even 60+ minutes~\cite{cordts16cvpr} per image depending on
the image size, complexity, and number of classes. 

To move from hundreds to thousands of classes we propose a new approach: given
a set of images with image-level labels, we build a weakly-supervised
model and use it to generate point-wise yes/no questions, e.g. ``is the point on a pumpkin?'' (fig.~\ref{fig:YesNo-UI}).
Such questions provide pointillism annotations that
we can use to train segmentation models and drive the generation of further questions.
This approach offers several benefits, outlined in the following paragraphs.

\paragraph{Minimal~input.}
Dense annotations are redundant since nearby pixels are correlated. Moreover, significant
time is spent at objects boundaries~\cite{caesar18cvpr} (that
could largely be automatically deduced), and for some classes the boundaries are
not well defined (e.g. human nose).
By asking about a specific point:
1)~we focus on the minimal supervision for semantic segmentation,
aiming to make the best use of human time;
2)~for a given annotation budget, we get more diverse annotations by collecting sparse
points over many images instead of dense annotations over few images;
3)~the point-wise annotation process is embarrassingly parallel, allowing many annotators to work simultaneously, even on the same image;
4)~it enables using active learning strategies to avoid repetitions across images.

\paragraph{Small~time~unit.}
For polygon annotations the minimal unit of annotation is an entire object,
which takes over a minute~\cite{lin14eccv}.
In our approach the unit is instead a single keystroke (or button press), thus shrinking unit time to a bare minimum (1s, \S\ref{subsec:Data-Stats}).
This is beneficial:
1) it allows annotators to start and stop work sessions quickly;
2) it enables to quickly assess a new annotator over diverse images to check whether they are familiar with a class %
(e.g. isopods vs insects);
3) it reduces the overall annotation effort when considering that multiple annotations per-point
might be needed to detect or disambiguate corner cases.

\paragraph{Incremental.}
Adding or improving existing annotations is done by
answering more questions of the same kind.
Unlike polygon drawing or brushing tools,
there is no ``erasing'' or ``editing'' procedure. 

\paragraph{Multi-class.}
Contrary to panoptic annotations~\cite{cordts16cvpr,caesar18cvpr,kirillov19cvpr},
with thousands of classes we cannot assume that every pixel has only
one class label (e.g. button $\leftrightarrow$ shirt $\leftrightarrow$ person).
We handle this by asking multiple classes per point, and collecting
their respective ``yes'' or ``no'' answers. With enough data collected,
one can also make statistics of classes likely or unlikely to co-occur
at the same point (e.g. honey \& spoon, versus honey \& toad) to guide
future questions.

\paragraph{No~guidelines.}
When scaling the number classes, defining
detailed per-class annotation instructions becomes a bottleneck~\cite{benenson19cvpr},
because a precise outline is requested (e.g. ``is the water part of the fountain?'').
Our yes/no questions are simpler, can
be immediately understood, and capture accepted/unacceptable namings of a
specific point. The fast annotations allows to aggregate multiple answers
per point to disambiguate difficult cases.

\paragraph{Access a large annotator pool.}
As our yes/no point-wise questions are extremely simple, the annotator can understand and answer them immediately, without undergoing any training nor understanding of the ultimate goal of such questions. The notion of object boundaries and segmentation masks are never exposed to the annotator. Because of this, {\em anyone} can annotate right away, thus opening up the potential to enrol a very large pool of annotators: anyone with an internet access.
Besides scaling up, such democratisation also enables to annotate a broader range of classes,
including classes known only in certain regions of the world (e.g.\ food ingredients in Taiwanese cuisine),
or only by people interested in niche areas (e.g.\ types of spiders).

\paragraph{Mobile-ready.}
The interface we propose is naturally suited for touchscreen
interactions, since we only ask to click one of three buttons (``yes'',
``no'', ``unsure''). In contrast, polygon annotations are most
comfortably done with a mouse, restricting annotators to a desk.
This mobile-ready design further extends the accessible pool of annotators.

In practice, scaling up annotations requires more than just reducing human annotation time.
All the mentioned benefits add up to make pointillism an attractive approach.

Note that any form of self-~\cite{misra20cvpr,chen20arxiv}, weak-~\cite{ahn19cvpr,wang20cvpr}, or web-supervision~\cite{divvala14lcvpr,jin17cvpr}
still needs additional direct pixel-level human supervision, either to adapt to the application
domain~\cite{mahajan18eccv},
to further improve quality,
or to evaluate the quality of the resulting models via a test set.
These methods reduce the volume of manually annotated data needed per class, but do not avoid the need to scale to many classes. Even in these scenarios large-scale data annotation campaigns are needed.

The rest of the paper is organized as follows.
Sec.~\S\ref{sec:Training-simulations} explains how to automatically select points to be annotated and assesses the quality of models trained from point annotations obtained in simulation.
Sec.~\S\ref{sec:Evaluating-models} explains how to evaluate model performance at test time based on point annotations only, and demonstrate that this leads to the same ranking of models as when using dense annotations.
Sec.~\S\ref{sec:human-annotators} switches to experiments with real human annotators and reports how fast and accurately they can answer our point questions.
Finally, sec.~\S\ref{sec:oidv7-point-annotations} describes a large-scale annotation campaign over \num{4000}+ classes on Open Images~\cite{kuznetsova20ijcv}.

\section{Points selection and model training}
\label{sec:Training-simulations}

An annotation campaign is defined by a set of questions.
The goal is to select diverse questions that
will be useful for training segmentation models.
For this we prefer questions that are likely to provide ``yes'' answers, as these provide a label for a point, whereas a ``no'' only partially constrains it.

A point question is defined by a triplet: image, point location $p$, and candidate class (e.g.\ ``pumpkin'' in fig.~\ref{fig:YesNo-UI}).
We inform the selection of this triplet using a weakly-supervised semantic segmentation
model $\mathcal{M}$ trained using only image-level labels (which we assume are given).
Unless otherwise specified, we use the state-of-the-art IRN method~\cite{ahn19cvpr} to construct $\mathcal{M}$.

\paragraph{Image selection.}
In our experiments we sample images uniformly out of the dataset to be annotated.
Depending on the application, one could prefer sampling images containing fewer classes, or using
$\mathcal{M}$ to select images with expected large areas of a class of interest.

\paragraph{Candidate class selection.}
Given a point $p$ the selection of the candidate class is informed
by the image-level labels and the model predictions scores $\mathcal{M}(p)$
at that point. We selected as candidate class the highest scoring one among those present in the image according to the image-level labels.
When running an annotation campaign, if the highest scoring class receives
a ``no'' answer, then the next highest scoring one is taken for the next question for point $p$.

\paragraph{Point selection.}
The points to be annotated should be diverse, cover well the classes of interest, and
be complementary to the information contained in $\mathcal{M}$.  
In \S\ref{subsec:Training-results} and \S\ref{subsec:COCOTas-results} we evaluate the effectiveness of training a semantic segmentation model from such points.

Once a first set of answers has been collected, we can update $\mathcal{M}$
by re-training it using the image-level labels and the point labels, and use
it to generate new, more precise, questions; thus feeding a virtuous
cycle of annotations.

Creating the experimental figures required overall the equivalent of \num{300} days of single-GPU training time.

\begin{figure*}
\begin{minipage}[t]{0.485\textwidth}%
\begin{centering}
\includegraphics[width=1\textwidth]{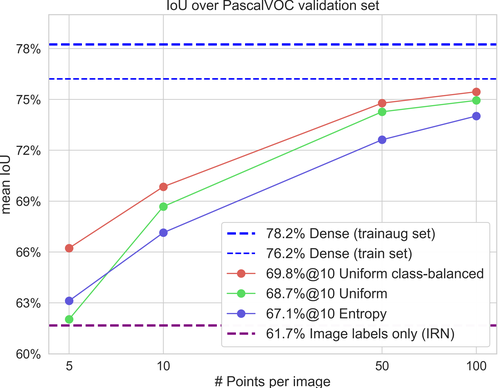}
\par\end{centering}
\caption{\label{fig:Pascal-Validation-set-comparison}Validation set comparison
of models trained from different point selection variants.}
\end{minipage}\hspace*{\fill}%
\begin{minipage}[t]{0.485\textwidth}%
\begin{centering}
\includegraphics[width=1.0\textwidth]{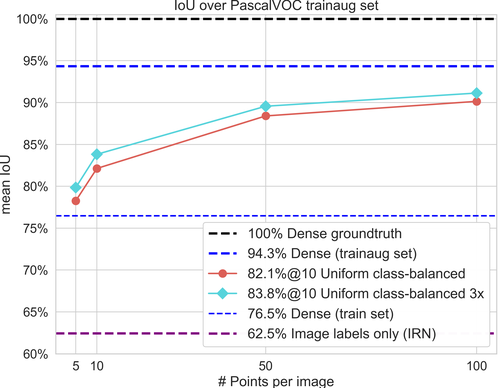}
\par\end{centering}
\caption{\label{fig:Pascal-training-set-comparison}Training set comparison
of models trained with different point selection variants.}
\end{minipage}
\end{figure*}

\begin{figure}
\begin{centering}
\hfill{}\includegraphics[width=0.65\columnwidth]{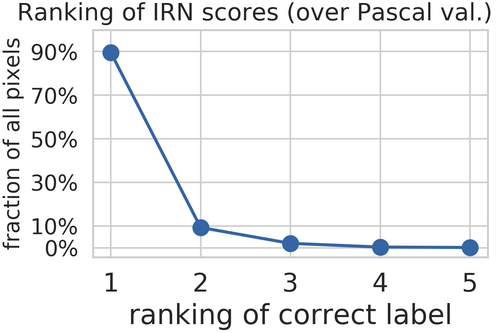}\hfill{}
\end{centering}
\caption{\label{fig:IRN-label-ranking}Ranking of label predictions from the
weakly-supervised $\mathcal{M}$ model (IRN). The correct label
is nearly always in the top-3. 
}
\end{figure}

\subsection{Experiments on PASCAL VOC}
\label{subsec:Training-results}

\paragraph{Training segmentation models from points.}
We use the PASCAL VOC~\cite{everingham15ijcv} semantic segmentation dataset
as our main experimental playground.
Our initial model $\mathcal{M}$ is a DeepLab V3 Xception65~\cite{chen18eccv} pre-trained for image classification on ILSVRC-12~\cite{russakovsky15ijcv} and then fine-tuned (300k steps) over the trainaug set ($10.6$k images~\cite{hariharan11iccv}) using pseudo-labels generated via IRN~\cite{ahn19cvpr} (weakly-supervised from image-level labels only).
Fig.~\ref{fig:Pascal-Validation-set-comparison} and~\ref{fig:Pascal-training-set-comparison}
show the results of training a DeepLab V3 Xception65~\cite{chen18eccv} model using labelled points, the equivalent of simulating a ``Yes'' answer for the correct label at each point.
We vary the number of points per image and the points selection strategy.
The training loss is computed only at the points with annotation, and the hyper-parameters are kept the same for all variants.
Fig.~\ref{fig:Pascal-Validation-set-comparison} shows models trained
over the PASCAL VOC trainaug set and evaluated over the validation set. 
In fig.~\ref{fig:Pascal-training-set-comparison} we evaluate the labels
reconstructed over the trainaug set: going from sparse to dense labels by running the trained model back over its training images. The ``3x'' tag denotes an ensemble of 3 DeepLab models trained over the same point labels.

\paragraph{Point selection.}
In fig.~\ref{fig:Pascal-Validation-set-comparison} we report the performance of models trained on points selected by different strategies.
For Pascal VOC, training models using points selected via %
simple spatial uniform sampling (`Uniform') performs better than %
uniform sampling in the per-image high entropy regions of the model $\mathcal{M}$ (`Entropy')~\cite{joshi09cvpr, mackowiak18bmvc}. %
The latter, in principle, focuses on difficult areas that are more informative than average. %
We also considered other point selection strategy variants (see appendix \S\ref{sec:Points-selection-strategies-details}), however these provided even worse results. %
Our best results come from the `Uniform class-balanced' strategy, %
which uses the pseudo-labels output from $\mathcal{M}$ to sample points in the same proportion for all classes. %
We also explored training from a combination of the pseudo-labels output from $\mathcal{M}$ 
and the point labels (e.g. including low-entropy regions of $\mathcal{M}$), but this did not improve results.
While $\mathcal{M}$ is useful to pick questions, it still makes
incorrect predictions for many pixels. %
Our point labels inject sparse but correct supervision, $\mathcal{M}$ injects denser but noisier supervision which ends up being more detrimental than helpful.%

\paragraph{Points per image.}
Fig.~\ref{fig:Pascal-Validation-set-comparison} reveals that labelling just 10 points per image already closes half of the performance gap between the weakly-supervised starting point (IRN model
$\mathcal{M}$), and the fully-supervised model ('Dense (trainaug set)'). With $50$ points per image, $80\%$ of the gap is closed. Overall, collecting 50 labeled points over 10.6k images provides
$95.7\%$ of the mIoU performance achievable by densely labelling them.%

Fig.~\ref{fig:Pascal-training-set-comparison} shows that labelling just 5 points per image leads to reconstructing the
ground-truth dense labelling at $80\%$ mean IoU, while with 50 points we can reach a pleasing $90\%$ (cyan curve).
We also evaluate the reconstruction ability of a model trained over the smaller training set (1.4k images) with dense ground-truth.
Interestingly, this reconstructs the trainaug set at only 76.5\% mIoU. %
This shows that it is better to have sparse samples covering the full domain, than dense samples on a small portion of it.

\vitto{possible cut, really I think all of it should go (especially if we wanna submit to CVPR'23)}
The results here serve as vanilla baselines, as indications of what is possible when using point annotations directly. %
For the sake of purity, all our methods in this paper use only
image-level labels and sparse point annotations. In practice one would consider
using transfer learning from another dataset with dense annotations when building $\mathcal{M}$ and the final
model, or use a mixed sparse and dense annotation strategy, similar to the
one discussed in~\cite{zlateski18cvpr}. Both of these would allow
to further improve results.

\paragraph{Pointwise class-ranking.}
Above we assumed a ``Yes'' answer for every question.
We evaluated the proportion of pixels where the correct class label is within the top-N classes predicted by the weakly-supervised $\mathcal{M}$ model (trained on trainaug, and tested on validation set, fig.~\ref{fig:IRN-label-ranking}).
With uniform spatial sampling, about 90\% of the points have their correct class as the top-1 prediction, and hence would require just one question to obtain a ``yes''. Only 10\% of the points would require 2 questions, and almost never more than 3 are required.
When using uniform class-balanced sampling the average number of questions to obtain a ``yes'' was
below 1.5 questions per point.
Hence, our process is efficient in the number of questions asked per point.

\begin{figure}
\begin{centering}
\includegraphics[width=1.0\columnwidth]{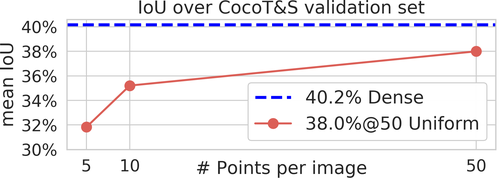}
\par\end{centering}
\caption{\label{fig:COCOTaS-validation-set}Model trained
over points sampled
on COCO T\&S. 
}
\end{figure}%

\subsection{Experiments on COCO Things \& Stuff}
\label{subsec:COCOTas-results}

We now move to the COCO Things \& Stuff dataset~\cite{lin14eccv,caesar18cvpr},
which also contains ``stuff'', i.e. background classes like grass and sky.
It offers 171 classes (80 things, 91 stuff). The training and validation set
have 118k and 5k images respectively. %
Fig.~\ref{fig:COCOTaS-validation-set} shows the results of training
a model using simulated "Yes" answers for spatially uniformly sampled points, using the same Xception65 DeepLab V3 model as before.
Like for PASCAL, with just 10 points
per image we recover $87\%$ of the mIoU performance of a fully supervised model, and with
50 points per image we get to $95\%$ (only two absolute percent points drop from 40\% mIoU to 38\%).
Just like for PASCAL, this indicates that what matters to train a model is
having a large-enough set of diverse points, rather than having spatially contiguous points (as in dense labelling).

\section{Evaluating models with points}
\label{sec:Evaluating-models}

Traditionally weakly-supervised methods still evaluate their models over
a test set with dense manual annotations~\cite{pathak15iccv,bearman16eccv,kolesnikov16eccv,cheng15cgf,papandreou15iccv,khoreva17cvpr}.
When scaling up annotations over thousands of classes, constructing an evaluation
set can by itself be daunting.
We show here that the proposed pointillism
annotations can also be used for evaluating semantic segmentation models.

The most common evaluation metric is intersection-over-union~\cite{everingham15ijcv} (IoU; a.k.a\ ``Jaccard Index'').
For a given class $c$, IoU for a test image is computed as the number of pixels that have class
$c$ in both the output of the model and the ground-truth segmentation map (intersection),
divided by the number of pixels which have class $c$ in either of them (union).
For evaluation over multiple test images, the common practice is to treat these as if concatenated into a large one~\cite{everingham15ijcv, lin14eccv, cordts16cvpr}.
The final number typically reported is the mean IoU over classes (mIoU).

We make the key observation that IoU operates over sets of pixels: it is insensitive to permutations
and to the local 2D structure. Therefore, we can evaluate IoU on a sparse
subset of test points with ground-truth labels.
This can be seen as a sampling approximation to the full IoU computed over dense annotations.
It remains to be determined how many points do we need to obtain a robust estimate of the full IoU value for a given class?
Even for datasets with only few thousands of images, the largest
classes would contain tens of millions of pixels.
However these pixels are very correlated and therefore redundant.
In this section we show experimentally that substantially fewer points (thousands instead of millions) suffice to estimate IoU robustly enough to evaluate (and rank) various segmentation methods.
To the best of our knowledge  this is the first work that exploits this property of the IoU measure, and explores it experimentally.

\begin{figure}
\begin{centering}
\includegraphics[width=1.0\columnwidth]{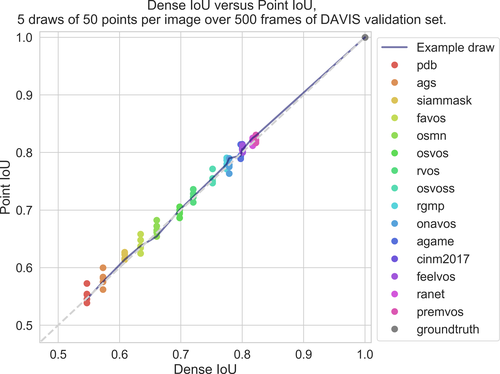}
\par\end{centering}
\caption{\label{fig:Eval-IoU-vs-point-IoU}Dense IoU versus point IoU,
5 draws of 50 points per image. Evaluating 15 segmentation methods (coloured dots) over 500 images. The point IoU estimate has low variance, and
the ranking across methods is well preserved.}
\end{figure}%

\paragraph{Experimental setup.}
To validate our proposed point IoU measure, we want to benchmark as many segmentation methods as possible.
We select the DAVIS 2017 video segmentation dataset~\cite{ponttuset17davis}, because its leaderboard provides the masks produced by many methods.
DAVIS contains challenging real-world videos for class-agnostic object segmentation, %
provides very accurate ground-truth annotations, features diverse types of objects, and is the main reference dataset in video object segmentation.
While DAVIS includes per-instance annotations, we disregard them and treat the task as a class-agnostic foreground versus background segmentation, matching the desired semantic segmentation task (there are no classes in DAVIS).
As reference for evaluation, we use the ranking of the 15 segmentation methods from the leaderboard according to dense IoU measured on the original ground-truth.

Fig.~\ref{fig:Eval-IoU-vs-point-IoU} plots the IoU of
the $15$ methods computed using the dense annotations or our points.
We evaluate on $500$ random frames out of the 2000 annotated ones in the $30$ videos
of the validation set.

For each frame we uniformly sample 50 points
(points per image - ppi) and get their ground-truth label from the 
dense ground-truth (simulating obtaining ``yes'' answers from annotators).
We evaluate the 15 segmentation methods either using the original dense annotations (dense IoU) or the sampled points (point IoU).
To study the statistical robustness of our method,
we repeat the sampling process 5 times (draws).

Fig.~\ref{fig:Eval-IoU-vs-point-IoU} shows the results of all 5 draws
and a line connecting the methods for one such draw.
A positive slope indicates that the ranking of methods by dense IoU is
preserved by point IoU.
The closer the line is to the diagonal, the closer are point IoU values to dense IoU values.
The average Kendall's $\tau$ rank correlation~\cite{kendall45biometrika}
over $5$ draws of $50$ points per image is $0.97$, showing that point
IoU preserves the ranking of methods very well.
As a reference, the average $\tau$ when evaluating dense IoU over the $15$
methods by sampling $5$ different sets of $500$ test frames is $0.94$.
Hence, the variation between dense IoU and point IoU is comparable to the variation in dense IoU among different test sets.
Additionally, point IoU leads to changes in ranking only for very closely performing methods, which have a mean delta dense IoU of $0.29\%$.
Finally, even with only 10 ppi, we already get $\tau$ 0.93, thus our proposed point IoU measure can be realised with very little annotation.

In principle, an IoU estimate will converge to a stable value given enough points.
In the setup above, over $500$ validation images, $25$k points ($50$ ppi) provide a near-perfect approximation to dense IoU, which instead requires annotating $153$M points.
For the multi-class scenario, we need to have enough sparse points per class (in
the order of $\sim10$k). Then the mean IoU across classes can be computed as usual.
The more diverse the evaluation points the better the measure of quality 
($1$ ppi over $10$k images is better than $10$k points over a single image).

\section{Pointillism with human annotators}
\label{sec:human-annotators}

In addition to the simulation experiments from \S\ref{sec:Training-simulations}, 
we also ran a campaign with real annotators over PASCAL VOC~\cite{everingham15ijcv} as a warm-up to the large scale campaign over Open Images~\cite{kuznetsova20ijcv} (\S\ref{sec:oidv7-point-annotations}).
We collected 10 points per image using the IRN model $\mathcal{M}$ with the uniform class-balanced sampling
strategy over PASCAL trainaug (see \S\ref{sec:Training-simulations}).  We instructed the annotators with
\textquotedbl simply answer based on your own beliefs. No need
to overthink the question.\textquotedbl.
We used the same interface and question formulation as in fig.~\ref{fig:YesNo-UI}.
We ordered the questions so as to present contiguous blocks of the same class to an annotator, 
to reduce context switching.
To reduce noise in the final point labels, we ask each question to 2-3 annotators, we call this `replication'.
On the first iteration, we consider a question answered if all its 2-3 replicas agree for a ``yes'' or ``no'' (left ``unresolved'' otherwise). We then do a second iteration on the unresolved questions, picking the second-highest scoring $\mathcal{M}(p)$ class as candidate.
After two iterations, we collected $572$k answers for $106$k points over $20$ classes.

\paragraph{Annotation time.}
The annotators answered each question in $0.8$ seconds (robust mean in $[10\%,\ 90\%]$ percentile range). Hence, this kind of questions are very fast to answer.

\paragraph{Comparison to polygon drawing.}
On average drawing polygons on one PASCAL image takes 216 seconds (2.7 objects~\cite{everingham15ijcv} times $80$ seconds per polygon~\cite{lin14eccv}).
Annotating $50$ points per image with our method involves answering on average $75$ yes/no questions, as some receive a `no' answer. At $0.8$ seconds per question this totals $60$ seconds (or $120$ seconds for $100$ points per image). As we have shown in fig.~\ref{fig:Pascal-Validation-set-comparison}, even just $50$ sparse points per image is already sufficient to reach $96$\% of the performance of a model trained on densely labelled images.
Hence, we conclude that our scheme is more efficient than traditional polygon drawing.

\paragraph{Quality.}
We observe $98$\% human-human agreement for questions with $3$ replica. This indicates that overall annotators have a consistent judgement even across very diverse classes.
Furthermore when comparing to the ground-truth annotations, $95$\% of all answers are correct (even $98$\% when considering only answers where all $3$ replica agree). Despite the annotators having only loose instructions, their answers match well the PASCAL annotations.
These results confirms that our questions are easy to answer and that human agreement is
a good proxy for correctness.%

\section{Point annotations on Open Images}
\label{sec:oidv7-point-annotations}

The Open Images dataset contains millions of images, along with image-level labels spanning $20$k classes.
When considering creating annotations for thousands of classes several challenges appear.
Even choosing a dictionary of classes to be annotated is non-trivial (e.g. not all classes can be localized, like ``night'').
Moreover, Open Images does not offer complete image-level labels, as at this scale this would have prohibitive cost on its own. This raises the question of how to select which labels we should annotate with points for each particular image.
Finally, methods such as IRN~\cite{ahn19cvpr} have not been shown to scale to a large number of classes, thus obtaining the initial weak segmentation model $\mathcal{M}$ needed to generate the point-wise yes-no questions is challenging too.

We address these challenges with a strategy based on the existing `Localized Narratives'~\cite{ponttuset20eccv} annotations (\S\ref{sec:localized-narratives}).
We then proceed to cover 
other images that do not have such annotations, with a second strategy based on a large weakly-supervised image-text model~\cite{jia2021icml} (\S\ref{sec:using-align-model}).

\subsection{Using localized narratives}
\label{sec:localized-narratives}

The Localized Narratives~\cite{ponttuset20eccv} (LN) annotations on Open Images are free-form spoken descriptions of an image produced by annotators while they simultaneously move their mouse over the regions they are describing. The voice recordings are then transcribed, and the resulting text remains temporally aligned with the mouse trace, thus providing approximate localization information for every word.
We process this data as follows, addressing the challenges listed above with an automated pipeline.

\paragraph{Dictionary of classes.}
First, 
we use commercial software %
to automatically extracts concepts from the LN captions and abstracts away the specific syntax or phrasings (e.g.\ ``aguacates'' $=$ ``avocado''). %
This tool automatically aggregates terms with near-identical semantics via named-entity disambiguation. %
This process triggers $54$k different entities (classes). %
Next, we filter the extracted entities to select only tangible, physical visual concepts (e.g.\ ``pelican'', as opposed to ``politics''). For this we use an existing knowledge base~\cite{noy19queue} which contains such tags. %
Finally, we intersect this set with an English dictionary to discard fine-grained classes (e.g.\ ``Nike Classic Cortez''), keeping only common concepts (e.g.\ ``sneakers'').
This leaves about $4$k visual concepts to be annotated.

\paragraph{Generating point questions.}
For each image, we aim to generate yes/no questions for the entities within our dictionary mentioned in its associated LN caption, which effectively provide (incomplete) image-level labels.

\begin{figure}
\begin{centering}
\includegraphics[width=1.0\columnwidth]{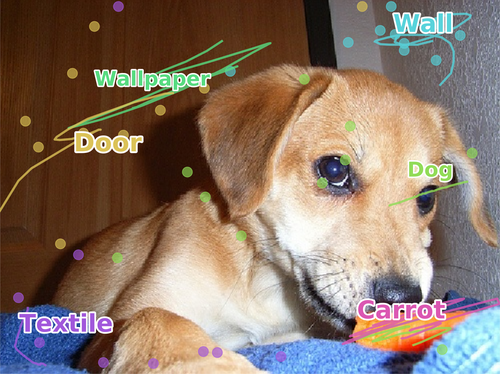}

\caption{\label{fig:Example-localized-narrative}Example Localized Narrative data; showing image,
detected entities, corresponding traces, and sampled question points (colour-coded by the class of the question that will be asked).}
\end{centering}
\end{figure}

To scale  $\mathcal{M}$ to thousands of classes, we propose to use here the mouse traces that come with the LNs.
Each word is associated with a mouse trace segment, providing a loose indication of its image location (fig.~\ref{fig:Example-localized-narrative}).
We first use these mouse traces to train a class-generic appearance embedding model. For this we use the penultimate layer of a segmentation model trained over mouse traces of the 200 most common classes%
.
This model can be used to derive an embedding for any pixel in any image.
We then build an image-specific $\mathcal{M}$ model, which outputs class scores for a pixel in that image via nearest-neighbour classification in embedding space. As class exemplar points (and their labels) we use the mouse trace segments of the image's LN.
This non-parametric classification process is much easier to train and flexible to deploy %
than a 4000-way classification model $\mathcal{M}$ on all Localized Narratives across all images. Also, the constraint imposed by the image-level labels are naturally preserved by our process.

Following the results of \S\ref{sec:Training-simulations}, for each image we sample $11$
points $p$ spatially uniformly, and select their candidate class as the top $\mathcal{M}(p)$ score.
This process forms the (image, point, class) triplets that define the questions to ask.
To reduce noise in the resulting annotations, we ask each question to $2{\sim}3$ annotators (`replication'). We also run multiple rounds of questions over the same point if it has only received "no" answers (see earlier fig.~\ref{fig:IRN-label-ranking} discussion), each round probing the next highest scoring $\mathcal{M}(p)$ class.

\subsection{Using an image-text model}
\label{sec:using-align-model}

The strategy described in \S\ref{sec:localized-narratives} requires per-image Localized Narrative annotations. To annotate images without them, we devise a second strategy based on the recent progress in webly-supervised image-text models~\cite{lu2019neurips, su2019iclr, li2019visualbert, li2020eccv, radford2021icml,jia2021icml}.

\begin{figure}
\begin{centering}
\includegraphics[width=0.498\columnwidth]{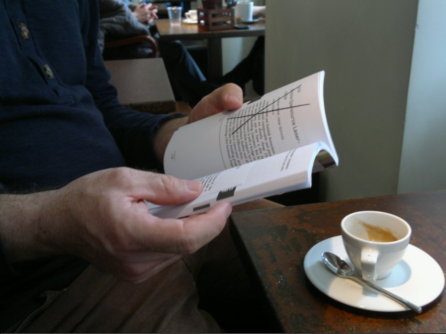}%
\hfill%
\includegraphics[width=0.498\columnwidth]{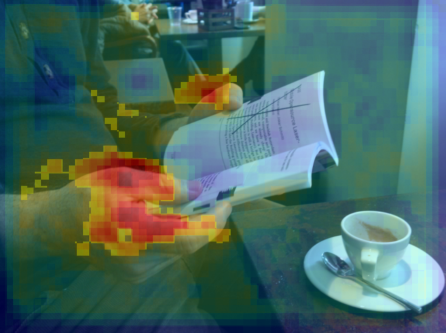}%
\\%
\includegraphics[width=0.498\columnwidth]{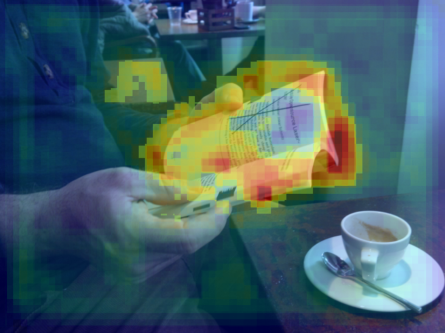}%
\hfill%
\includegraphics[width=0.498\columnwidth]{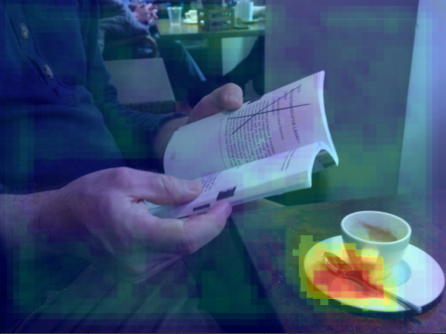}%
\caption{\label{fig:weak-segmentation-examples}Example of noisy zero-shot segmentation from the image-text model~\cite{jia2021icml}. Input image, classes hand, book, and spoon.}
\end{centering}
\end{figure}

These models are trained to embed images and their captions into the same space using large volumes of noisy web data. It has been shown that these image-text encoder models can be used as effective zero-shot classifiers, and that their internal representation captures reasonably well the spatial extent of visual concepts in an image~\cite{caron2021iccv, ghiasi2021arxiv, zabari2021arxiv, luddecke2022cvpr, ding2022cvpr, wang2022cvpr}. We thus modify the outputs one such model~\cite{jia2021icml} to generate noisy zero-shot semantic segmentation (fig.~\ref{fig:weak-segmentation-examples}). For the questions proposal task, accurate segmentation is not necessary (since point annotations around the boundaries are informative enough).

To obtain the image-level labels for each image, we query the model as a zero-shot classifier over each class in the dictionary identified in~\S\ref{sec:localized-narratives}.
For classes with high-enough score, we then run the model as a zero-shot segmenter to obtain rough per-class masks, thus using the image-text model as class-generic $\mathcal{M}$. %
The (image, point, class) triplets are formed by sampling superpixel centroid points inside and around the boundaries of the generated masks and selecting the per-point highest scoring class $\mathcal{M}(p)$ as question class to send to annotators. %
The multiple rounds of question answering are identical to the Localized Narratives \S\ref{sec:localized-narratives} and PASCAL VOC \S\ref{sec:human-annotators} cases.

\begin{figure}
  \begin{centering}
\includegraphics[width=1.0\columnwidth]{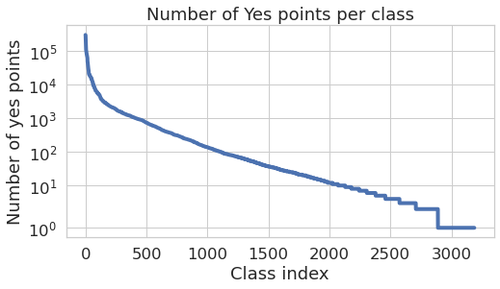}
\caption{\label{fig:Yes-points-per-class}Number of points with yes answer, per class; the curve follows Zipf's law, where many classes have few yes points.}
  \end{centering}
\end{figure}

\begin{figure}
\begin{centering}
\includegraphics[width=0.498\columnwidth]{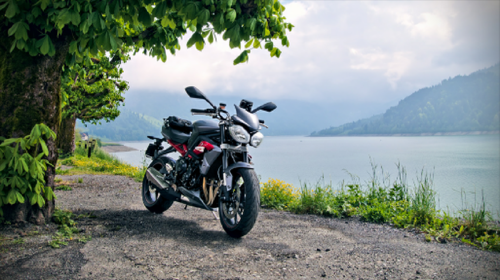}%
\hfill
\includegraphics[width=0.498\columnwidth]{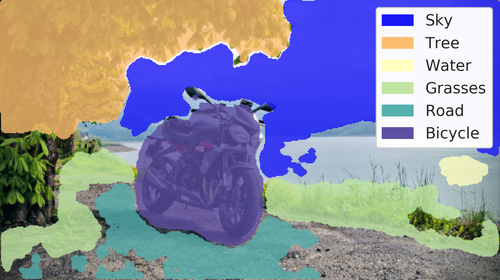}%
\\
\includegraphics[width=0.498\columnwidth]{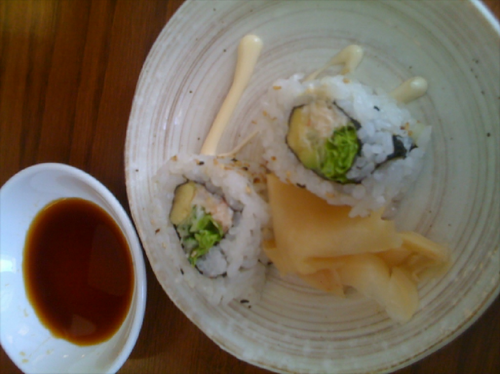}%
\hfill
\includegraphics[width=0.498\columnwidth]{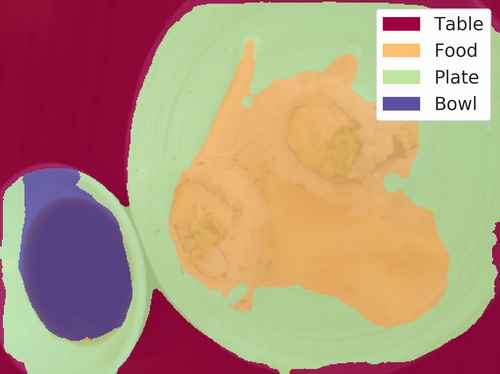}%
\caption{\label{fig:Oidv5Model-labels}Example outputs from segmentation mode trained with points from 235 Open Images classes.}
\end{centering}
\end{figure}%

\subsection{Data statistics}
\label{subsec:Data-Stats}

\paragraph{Total answers.}
After multiple iterations of annotation, we sent in total 65.2M questions over 692k images.
These questions cover 13.0M unique points, and lead to 22.6M point-labels over \num{4171} classes ($1.7$ average labels per point, $2.9$ average answers per label).
The $22.6$M labels contain $4.2$M Yes, $16.1$M No, and $2.3$M Unresolved.
The negative (No) labels are useful for training models. Keep in mind that the absence of positive point labels for a class does not mean that the class is absent in an image (it could be present in a region not covered by any annotated point). To train semantic models it is thus necessary to use reliable negative labels, like the existing negative image-level labels in Open Images, and from the new point-level ones. Unresolved points with multiple answers inform of areas where the class is ambiguous even for humans. %
Fig.~\ref{fig:yes-no-examples} and appendix \S\ref{sec:example_yes_no_points} show example images with the collected Yes and No labels (see also the online visualiser on the \href{https://g.co/dataset/open-images}{Open Images website}).

Out of the \num{4171} classes, \num{3189} have at least one Yes point label, and \num{2664} have $\geq 3$.
Fig.~\ref{fig:Yes-points-per-class} shows the distribution of Yes answers per class, which follows Zipf's  law, as expected.
Examples of classes with high number of Yes points are ``wood'', ``cloud'', ``glass''; with a mid number are ``milkshake'', ``telescope'', ``yak''; and with a low number ``drumhead'', ``paratha'', ``wrinkle''.

\paragraph{Time \& Quality.}
On Open Images, the annotators answered each question slightly slower than in PASCAL, this is expected since the images are more complex and the classes more diverse. On average each answer took $1.1$s (robust mean), which is still very fast.
Human-human agreement remains similar to the PASCAL case, with 98\% agreement for questions with 3 %
answers.

Training effectively a segmentation model over more than \num{4000} classes remains an open problem. As a proof of concept, we trained a model over 235 classes of interest and we show some example outputs in fig.~\ref{fig:Oidv5Model-labels}.

\paragraph{Open Images V7 release.}
The latest release of Open Images introduces point labels as a new annotation type on top of the existing image labels, bounding boxes, visual relations, instance masks, and localized narratives.
The V7 release includes the $22.6$M point labels over $4.2$k classes discussed above, as well as $1.5$M point labels over $4.1$k classes from~\cite{ponttuset2019arxiv}, and $42.3$M point labels over \num{389} classes, by-products of the interactive segmentation process of~\cite{benenson19cvpr}.
(The method from~\cite{ponttuset2019arxiv} provides diverse annotations, but is significantly slower than the approach presented here since it requires text typing for each point.
The method from~\cite{benenson19cvpr} is closer in speed, but has limited scalability over classes since it requires bounding boxes or a pre-existing segmentation model for the classes of interest, and it is not suitable for annotations on  mobile.)
\vitto{agreed to keep this last point for arXiv, skip it for CVPR'23 submission (IF we submit).}

After data conversion and merging, the V7 release includes a total of $66$M point-level labels over \num{5827} classes spanning $1.4$M images (\num{2033} classes appearing in train+val+test set, \num{5180} classes with >=1 Yes points, \num{3739} classes >= 3 Yes points).
For more information about this data, please consult the \href{https://g.co/dataset/open-images}{Open Images website}.

\section{Conclusions}
\label{sec:Conclusions}

As the scale of problems tackled by computer vision grows, the data
annotation problem becomes more and more pressing.
We have discussed how traditional semantic
segmentation annotation approaches fall short when considering growing
at scale, and have proposed a new pointillism approach, characterised by its extremely
low entry barrier, minimal annotation-time quantum, being mobile friendly,
and being embarrassingly parallel.
Our experimental results show that these annotations are fast, reliable,
and enable adequate model training and evaluation. The proposed
approach relies on weak-supervision and active-learning methods and
establish a virtuous cycle were stronger models enable more efficient
annotations. %

\vitto{possible cut}
Just like the algorithmic models have evolved in the last decade,
it is time to also evolve our algorithmic approach to data annotation.

\paragraph{Acknowledgements.} The authors would like to thank Jordi Pont-Tuset for the support provided when preparing this manuscript and the Open Images V7 release.

{\small
\bibliographystyle{ieee_fullname}
\bibliography{shortstrings,loco}
}

\clearpage 
\appendix

\section{Example Yes/No points}
\label{sec:example_yes_no_points}

\begin{figure*}
    \centering

\includegraphics[width=0.9\columnwidth, trim=0 -1.5mm 0 0]{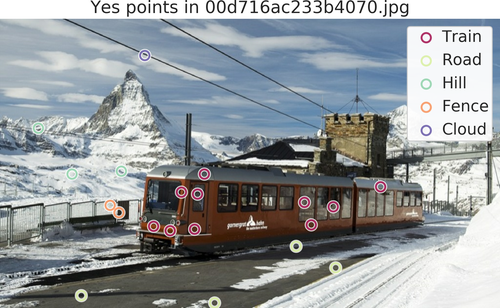}\hspace{2mm}
\includegraphics[width=0.9\columnwidth]{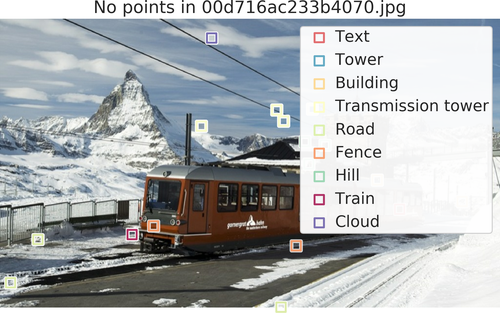} \\\vspace{0.5em}
    
    \includegraphics[width=0.9\columnwidth, trim=0 -1.0mm 0 0]{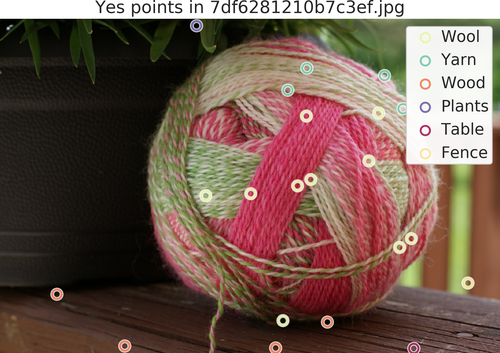}\hspace{2mm}
    \includegraphics[width=0.9\columnwidth]{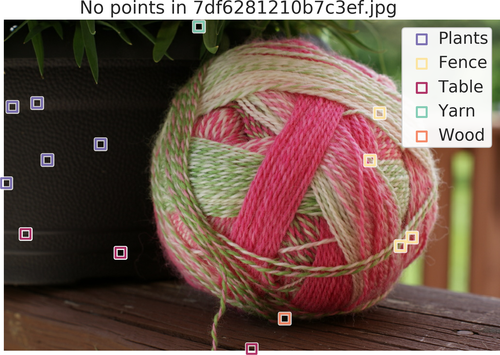}\\\vspace{0.5em}
    
    \includegraphics[width=0.9\columnwidth, trim=0 -1.0mm 0 0]{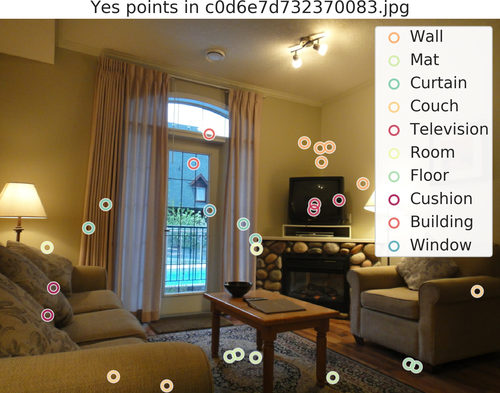}\hspace{2mm}
    \includegraphics[width=0.9\columnwidth]{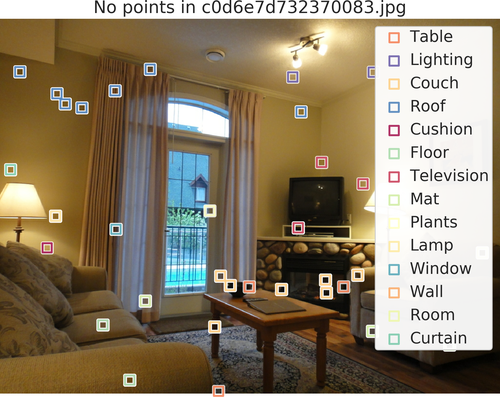}\\\vspace{0.5em}

    \includegraphics[width=0.9\columnwidth]{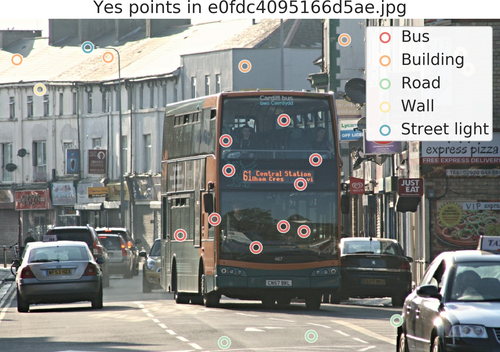}\hspace{2mm}
    \includegraphics[width=0.9\columnwidth]{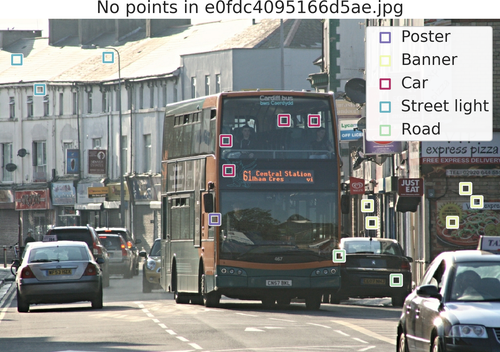}\\\vspace{0.5em}
    
    \caption{Example collected yes/no point labels. Circles indicate ``yes'' labels, and squares ``no'' labels.}
    \label{fig:Example-yes-no-points-a}
\end{figure*}

\begin{figure*}
    \centering

    \includegraphics[width=0.9\columnwidth]{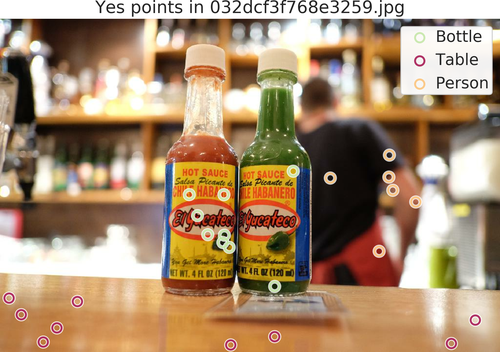}\hspace{2mm}
    \includegraphics[width=0.9\columnwidth]{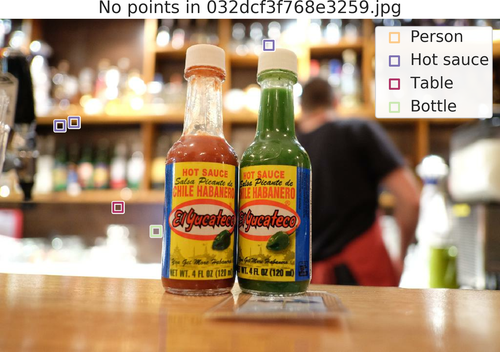}\\\vspace{0.5em}

    \includegraphics[width=0.9\columnwidth]{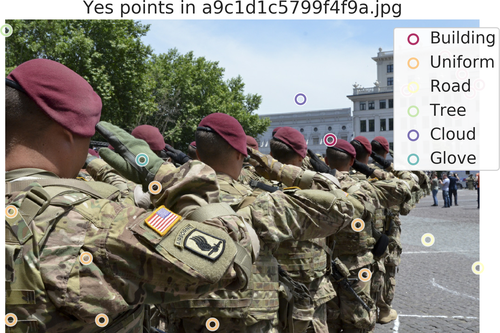}\hspace{2mm}
    \includegraphics[width=0.9\columnwidth]{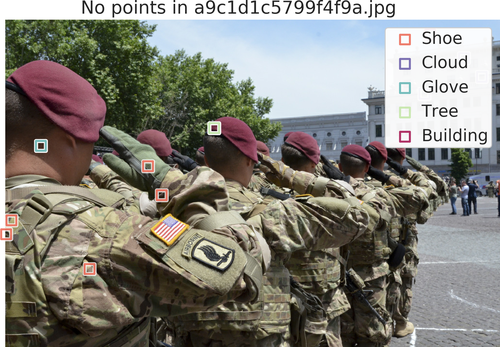}\\\vspace{0.5em}
    
    \includegraphics[width=0.9\columnwidth]{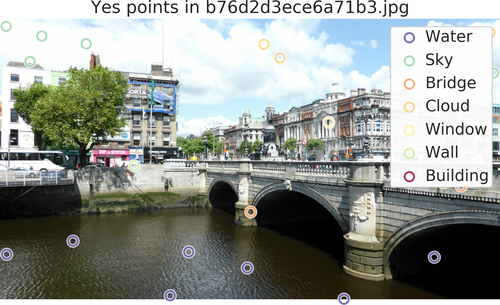}\hspace{2mm}
    \includegraphics[width=0.9\columnwidth]{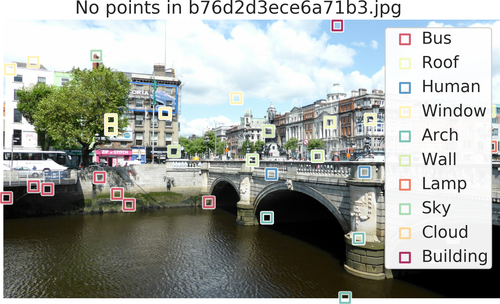}\\\vspace{0.5em}
    
    \includegraphics[width=0.9\columnwidth]{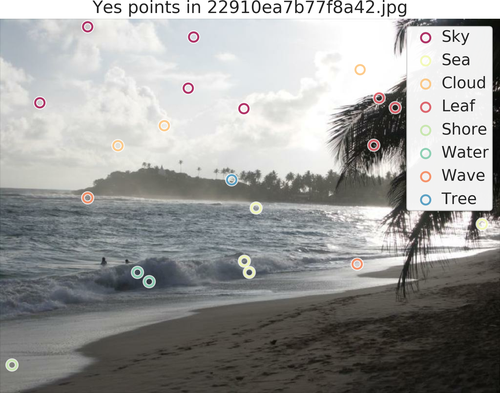}\hspace{2mm}
    \includegraphics[width=0.9\columnwidth]{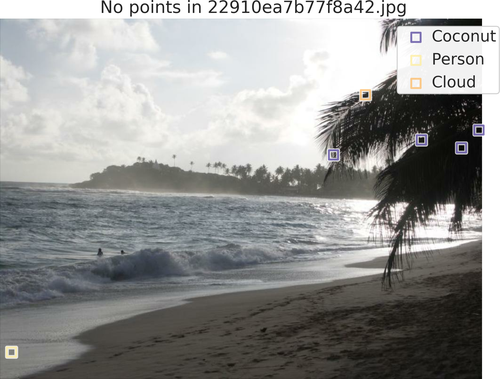}\\\vspace{0.5em}

    \caption{Example collected yes/no point labels. Circles indicate ``yes'' labels, and squares ``no'' labels.}
    \label{fig:Example-yes-no-points-b}
\end{figure*}

\begin{figure*}
    \centering

    \includegraphics[width=0.9\columnwidth]{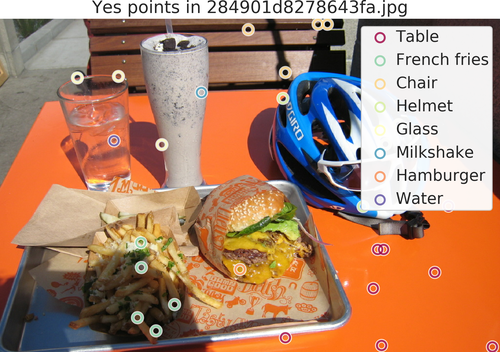}\hspace{2mm}
    \includegraphics[width=0.9\columnwidth]{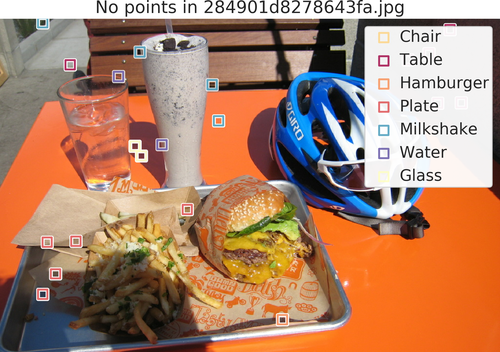}\\\vspace{0.5em}

    \includegraphics[width=0.9\columnwidth]{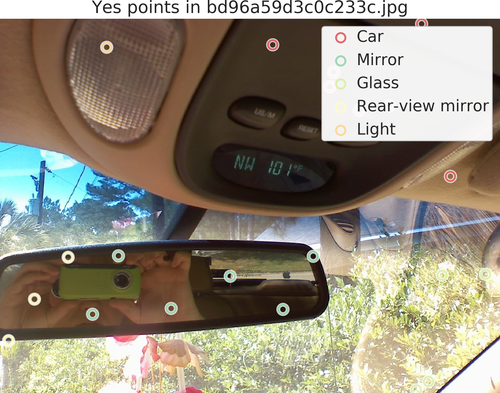}\hspace{2mm}
    \includegraphics[width=0.9\columnwidth]{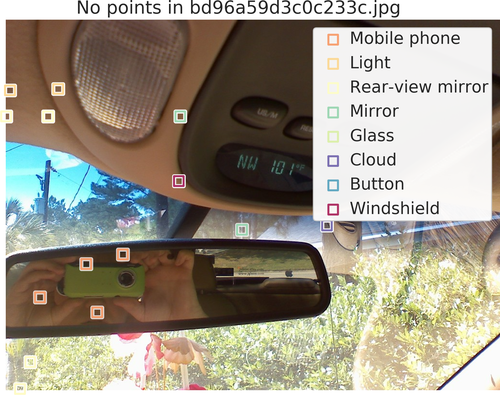}\\\vspace{0.5em}
    
    \includegraphics[width=0.47\columnwidth]{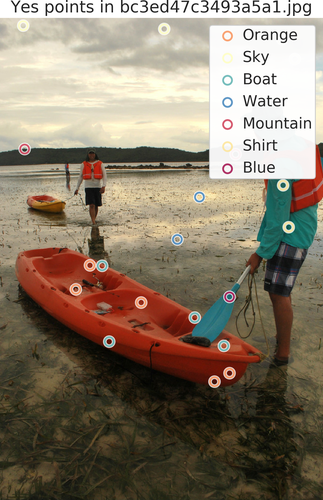}\hspace{1mm}
    \includegraphics[width=0.47\columnwidth]{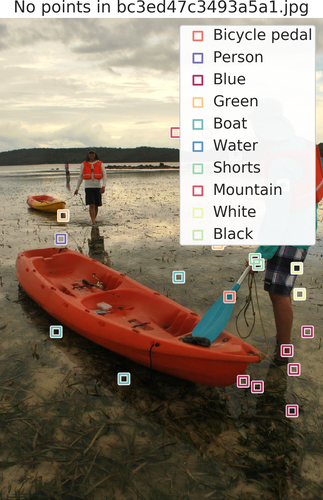}\hspace{2mm}
    \includegraphics[width=0.47\columnwidth]{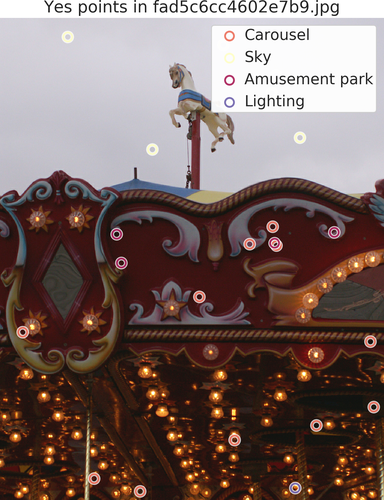}\hspace{1mm}
    \includegraphics[width=0.47\columnwidth]{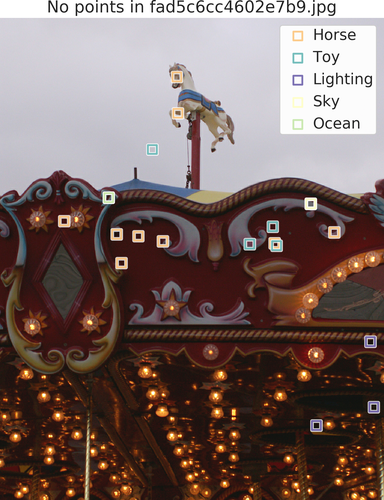}\\\vspace{0.5em}
    
    \caption{Example collected yes/no point labels. Circles indicate ``yes'' labels, and squares ``no'' labels.}
    \label{fig:Example-yes-no-points-c}
\end{figure*}

\begin{figure*}
    \centering

    \includegraphics[width=0.85\columnwidth]{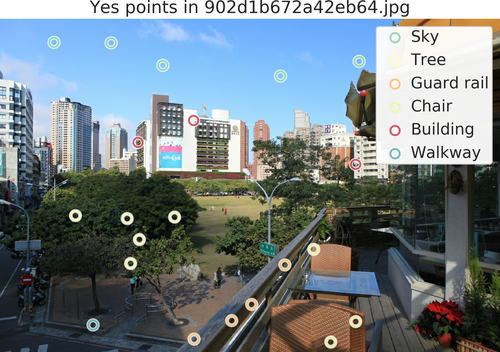}\hspace{2mm}
    \includegraphics[width=0.85\columnwidth]{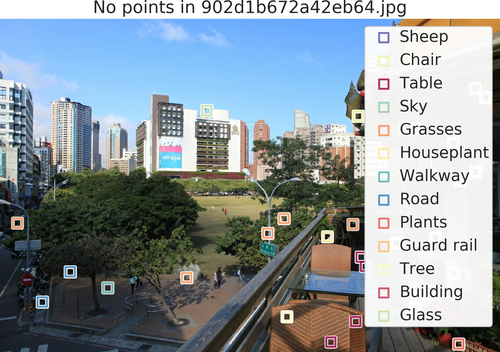}\\\vspace{0.5em}
    
    \includegraphics[width=0.79\columnwidth]{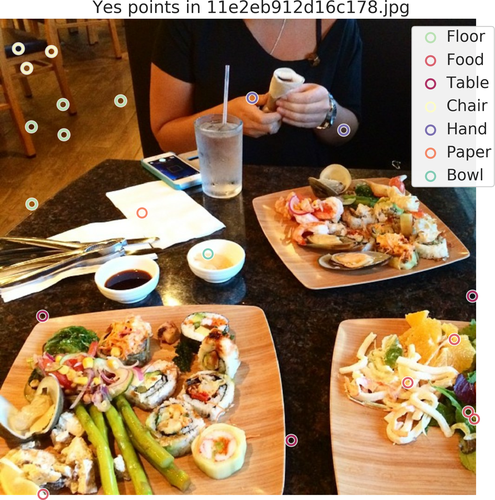}\hspace{2mm}
    \includegraphics[width=0.8\columnwidth]{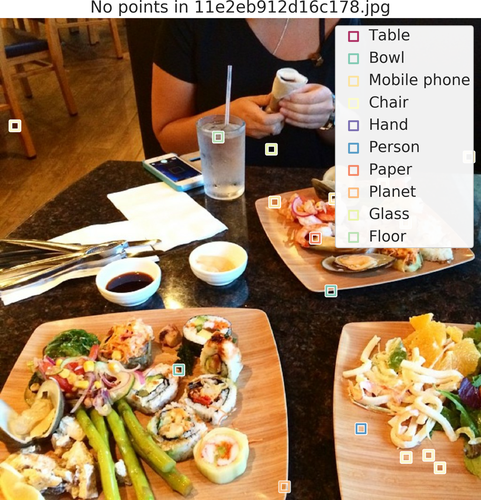}\\\vspace{0.5em}

    \includegraphics[width=0.85\columnwidth]{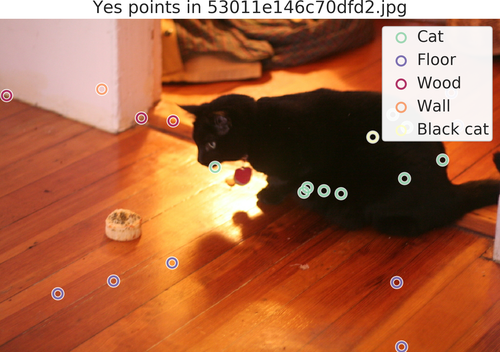}\hspace{2mm}
    \includegraphics[width=0.85\columnwidth]{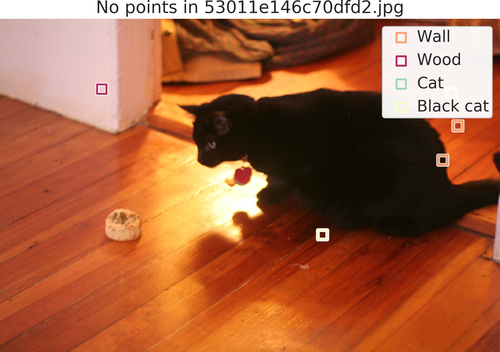}\\\vspace{0.5em}
    
    \includegraphics[width=0.85\columnwidth]{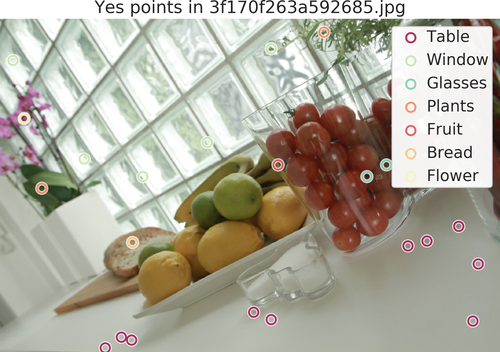}\hspace{2mm}
    \includegraphics[width=0.85\columnwidth]{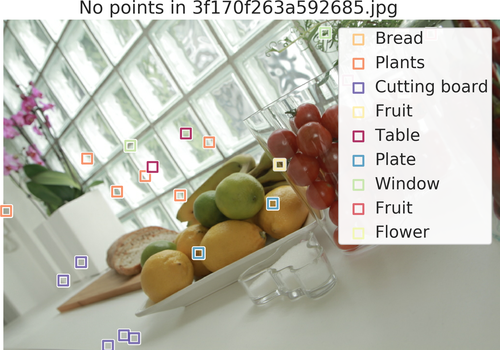}\\\vspace{0.5em}
    
    \caption{Example collected yes/no point labels. Circles indicate ``yes'' labels, and squares ``no'' labels.}
    \label{fig:Example-yes-no-points-d}
\end{figure*}

\vspace{-0.25em}
Figures~\ref{fig:Example-yes-no-points-a}, \ref{fig:Example-yes-no-points-b}, \ref{fig:Example-yes-no-points-c} and \ref{fig:Example-yes-no-points-d} present example yes/no answers collected over the Open Image training images.
Additional examples available in \href{https://g.co/dataset/open-images}{the online visualiser}.

\section{Point selection strategies}
\label{sec:Points-selection-strategies-details}

\vspace{-0.25em}
Section~\ref{sec:Training-simulations} presented results for how the point selection strategy affects the quality of a model trained on those points. Here we include some preliminary experiments done before training any model, that compare usual active learning data sampling strategies.

\vspace{-0.25em}
\paragraph{Experimental setup.}
We use the PASCAL VOC~\cite{everingham15ijcv} semantic segmentation dataset
as our experimental playground. The original training set contains $1.4$k
images, the augmented training set of~\cite{hariharan11iccv} has $10.6$k
images (trainaug), and the validation set has $1.4$k images.
We run the weakly-supervised IRN technique~\cite{ahn19cvpr} on trainaug to produce pseudo-labels at each pixel, starting from image-level labels only.
Our initial model $\mathcal{M}$ is a DeepLab V3 Xception65~\cite{chen18eccv} pre-trained for image classification on ILSVRC-12~\cite{russakovsky15ijcv} and then fine-tuned ($300$k steps) over the trainaug set using these pseudo-labels.
Thus the overall process to build $\mathcal{M}$ uses only image-level labels.

\begin{figure}
\begin{centering}
\includegraphics[width=0.85\columnwidth]{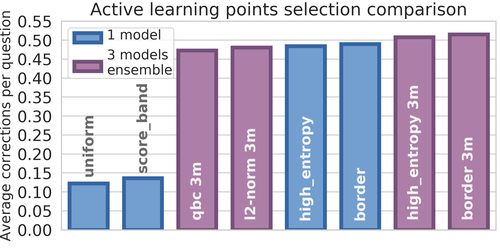}
\par\end{centering}
\vspace{-0.5em}
\vspace{-0.25em}
\caption{\label{fig:Active-learning-comparison}Comparison of active learning
strategies for selecting points. Vertical axis indicates the fraction
of questions complementary to model $\mathcal{M}$. Simply sampling nearby the semantic boundaries
is as effective as classical active learning methods.}
\vspace{-1em}
\end{figure}

\vspace{-0.25em}
\paragraph{Point selection.}
Fig.~\ref{fig:Active-learning-comparison} compares several active learning strategies to select points to be annotated (10 points per image over the validation set).
As active learning aims at finding the most informative points for $\mathcal{M}$, we evaluate which fraction of the selected points has a ground-truth label different than the highest scoring one predicted by $\mathcal{M}$. These points are expected to more informative to train a model, since they are closer to the decision boundaries.

Each selection method we consider defines a different subset of all points in an image, and then samples uniformly within it. Hence, we specify a method concisely by the subset it defines:
($\mathtt{uniform}$) all image points; 
($\mathtt{score\_band}$) points with predicted score $\in[0.8,\,0.9]$;
($\mathtt{border}$) points on the boundaries between predicted semantic class regions;
($\mathtt{high\_entropy}$) top-1\% high-entropy~\cite{joshi09cvpr, mackowiak18bmvc} points in the image (%
entropy of the distribution over classes output by the model);
($\mathtt{l2-norm\ 3m}$) top-1\% points with highest $l^{2}$-norm between three DeepLab models trained from the same IRN predictions (ensemble disagreement);
($\mathtt{qbc\ 3m}$) top-1\% points according to the query-by-committee measure~\cite{cohn94ml} (based on Jensen-Shannon divergence).
We also include $\mathtt{high\_entropy\ 3m}$ and $\mathtt{border\ 3m}$, computed over the average score of an ensemble of 3 DeepLab models.

Fig.~\ref{fig:Active-learning-comparison} shows
that the classical active learning methods (like $\mathtt{high\_entropy}$ and $\mathtt{qbc}$) are much better than uniform sampling at finding errors in the $\mathcal{M}$ predictions.
However they are not better that the domain-specific heuristic of sampling along the semantic borders. In practice semantic
segmentation models also deliver predictions with high class entropy along these borders,
which is why $\mathtt{high\_entropy}$ and $\mathtt{border}$ perform
similarly. %
The ensembles provide a small incremental gain over single model results.

The results of section \S\ref{subsec:Training-results} show that the usual active learning proxy metric of "fraction of points where $\mathcal{M}$ is confused", does not translate well to the performance (mean IoU) of models trained on those points (which is what truly matters).
We see there that in this metric simple methods like uniform sampling outperform high-entropy sampling.
We also tried there methods like query-by-committee sampling and got even worse results. It seems that for semantic segmentation it is most useful to focus on well distributed training examples, without much concern on collecting samples near the decision boundaries.

\section{Evaluating models with points}

\begin{figure}[h!]
    \vspace{-0.5em}
    \centering
    \includegraphics[width=0.9\columnwidth]{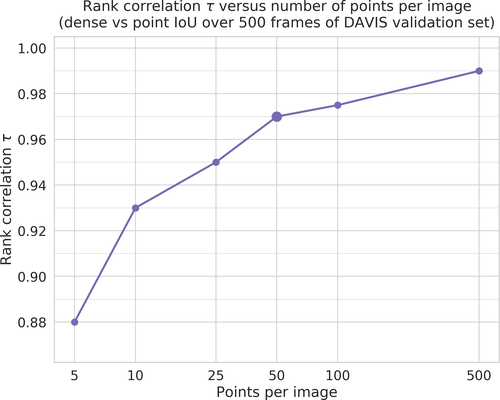}
    \caption{Kendall's rank correlation coefficient $\tau$ (between dense and point IoU) when using different number of points per image.}
    \vspace{-0.5em}
    \label{fig:tau-vs-ppi}
    \vspace{-1em}
\end{figure}

\vspace{-0.25em}
Fig.~\ref{fig:tau-vs-ppi} complements the results of sec.~\ref{sec:Evaluating-models}. %
It shows how the rank correlation changes as we vary the number of points per image. %
Each point represents the average Kendall's $\tau$ rank correlation coefficient~\cite{kendall45biometrika} between dense IoU and point IoU over 5 draws of the same 15 segmentation methods from fig.~\ref{fig:Eval-IoU-vs-point-IoU}. %
Fig.~\ref{fig:tau-vs-ppi} shows that for the scenario at hand the point IoU remains a good proxy for dense IoU even for as few as 10 ground truth points per image.

\end{document}